\documentclass{amia}
\usepackage{graphicx}
\usepackage[labelfont=bf]{caption}
\usepackage[superscript,nomove]{cite}
\usepackage{color}
\usepackage{url}
\usepackage{amsmath}
\usepackage{amsfonts}
\usepackage{amssymb}
\usepackage{booktabs}
\usepackage{wrapfig}
\usepackage{ulem}
\usepackage{tabularx}
\usepackage{enumitem}
\usepackage{comment}

\begin{document}

\title{Prior Knowledge Enhances Radiology Report Generation}

\author{
    Song Wang$^{1}$, Liyan Tang$^{1}$, Mingquan Lin, Ph.D.$^{2}$, George Shih, M.D.$^3$, Ying Ding, Ph.D.$^{1}$, Yifan Peng, Ph.D.$^{2}$
}

\institutes{$^{1}$ The University of Texas at Austin, Austin, TX, USA; $^{2}$ Population Health Sciences, Weill Cornell Medicine, New York, NY, USA; $^{3}$ Department of Radiology, Weill Cornell Medicine, New York, NY, USA
}

\maketitle

\noindent{\bf Abstract}

\textit{Radiology report generation aims to produce computer-aided diagnoses to alleviate the workload of radiologists and has drawn increasing attention recently. However, previous deep learning methods tend to neglect the mutual influences between medical findings, which can be the bottleneck that limits the quality of generated reports. In this work, we propose to mine and represent the associations among medical findings in an informative knowledge graph and incorporate this prior knowledge with radiology report generation to help improve the quality of generated reports. Experiment results demonstrate the superior performance of our proposed method on the IU X-ray dataset with a ROUGE-L of 0.384$\pm$0.007 and CIDEr of 0.340$\pm$0.011. Compared with previous works, our model achieves an average of 1.6\% improvement (2.0\% and 1.5\% improvements in CIDEr and  ROUGE-L, respectively). The experiments suggest that prior knowledge can bring performance gains to accurate radiology report generation. We will make the code publicly available at \url{https://github.com/bionlplab/report_generation_amia2022}.}

\section{Introduction}

A radiology report provides a translation of radiographs into text, presenting a synopsis of the process of detailed findings and thoughtful impressions\cite{pahadia_khurana_geha_deahl_2020}. 
It gives descriptive information about a patient's history, symptoms, and interpretations of relevant radiology images\cite{shin2015interleaved}. 
Therefore, radiology report writing has long been a time-consuming and labor-some process that requires domain expertise. 
To alleviate the burden of radiologists, there is an unmet need to develop automatic report generation systems. 

\begin{wrapfigure}{r}{0.65\columnwidth}
    \centering
    \vspace{-1em}
    \includegraphics[width=.65\columnwidth,page=2,clip,trim=0 24em 41em 0em]{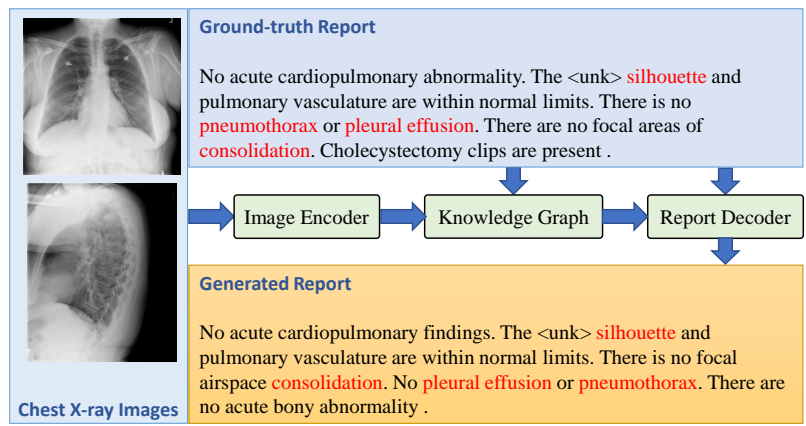}
    \caption{An example of the generated report. Concepts marked in red are the concept nodes in our knowledge graph. Chest X-ray images are encoded first, the image representations will then be used to generate texts.}
    \label{fig:eyecatcher}
    \vspace{.5em}
\end{wrapfigure}

Radiology report generation takes chest X-ray images as input and generates a descriptive report to support better diagnostic conclusion inferences beyond disease labels. This task is close to visual captioning\cite{rennie2017selfcritical,lu2017knowing,anderson2018bottomup, agrawal2016vqa} but posts new challenges. First, radiology report generation outputs a sequence of sentences, while visual captioning usually produces only one sentence. Second, radiology report generation requires extensive domain knowledge to produce clinical-coherent texts. For example, it must follow critical protocols, including the correct use of medical terms to describe normal and abnormal medical observations\cite{shin2015interleaved}.

Radiology report generation has attracted more attention recently\cite{jing-2018,wang2018tienet,jing-etal-2019-show,chen-etal-2020-generating-radiology, xue2018, liu2019clinically}. Xue et al. fused the visual features and the semantic features of the last sentence through an attention mechanism, and used this fusion to generate the next sentence in a recurrent generative manner\cite{xue2018} . Wang et al. presented a text-image embedding network to jointly learn the text and image information, and integrated an end-to-end trainable CNN-LSTM architecture with multi-level attention models for chest X-ray reporting\cite{wang2018tienet}. Jing et al. employed a co-attention model over visual features and textual embeddings, and proposed a multi-task learning framework that jointly predicts disease tags and generates reports\cite{jing-2018} . Chen et al. recorded vital information during the generation process by presenting a memory-driven Transformer to further assist report generation\cite{chen-etal-2020-generating-radiology} . Jing et al. used reinforcement learning to exploit the structure information between and within report sections for generating high-quality reports\cite{jing-etal-2019-show}. Liu et al. combined self-critical sequence training with reinforcement learning to optimize the correct mentions of disease keywords in reporting\cite{liu2019clinically}.

Though achieving good results, few works considered incorporating prior knowledge, which can provide supplementary information for accurate reporting. For example, medical observations presented in a chest X-ray image are usually not isolated from each other, where underlying mutual influences may exist. Compared to experienced radiologists aware of such relationships, deep learning methods tend to suffer from the lack of knowledge if not explicitly taught, which limits the generation accuracy. Modeling the associations among medical observations in the form of a knowledge graph enables us to further utilize prior knowledge to produce high-quality reports. To this end, Zhang et al.\cite{zhang_when_2020} and Li et al.\cite{li2019knowledgedriven} combined graph-based knowledge inference with an encoder-decoder pipeline for radiology report generation. However, their prior knowledge is manually pre-defined, thus, requires domain experts to be closely involved in the design and implementation of the system. Owing to the rigid graph, their approaches usually achieve a high precision but may miss some important findings. While it is feasible to manually identify and implement a high-quality knowledge graph to achieve good precision, it is often impractical to exhaustively encode all the nodes and relations in this manner. Our work enables the text-mined prior knowledge as a universal knowledge graph to mitigate some of these concerns.

This paper presents an innovative framework of knowledge-based report generation, which seamlessly integrates prior domain and linguistic knowledge at different levels. First, we study a data-driven approach to automatically capture the intrinsic associations among the concepts in the RadLex radiology ontology \cite{langlotz_radlex_2006}. This prior knowledge serves as a natural extension to the human-designed one \cite{zhang_when_2020}. Disease findings are defined as nodes in the graph, and correlated findings are connected to influence each other during graph propagation. Second, we build a graph convolutional neural network to model the prior knowledge on chest findings\cite{yao_exploring_2018}. Frontal-view and lateral-view images of chest X-ray are fed into a convolutional neural network extractor for image feature extraction, and the features together with the built graph are passed to a three-layer graph convolution network through an attention mechanism to learn dedicated features for each graph node. Later these node features are passed to two branches, one linear classifier for disease classification, and one two-level decoder for report generation. Different from previous studies, extra text-mined concepts are included in the model as auxiliary nodes so that the model enriches its expression power by training on existing datasets with image-level diseases annotated. We hypothesize that these text-mined labels may reflect known features identified in chest X-rays and add granularity to the association strength of those features in the generated reports. 

To train the model, we adopt an existing two-step procedure \cite{zhang_when_2020}: multi-label classification followed by report generation. Such a training strategy simulates the reading routine of radiologists by first observing multiple findings when they read medical images and then compiling radiological reports. Specifically, we first train a multi-label classifier where each class label corresponds to one medical finding and hence one node in the knowledge graph. After training the multi-label classifier, we keep the classifier frozen, and train a two-level decoder that consists of one topic-level Long Short-Term Memory (LSTM\cite{HochSchm-1997}) and one word-level LSTM. The two-level decoder encourages each generated sentence to focus on one different topic. Fig \ref{fig:eyecatcher} shows the generation pipeline and an example of the generated report.

Our contributions are outlined as follows. (1) We text-mine and model the prior knowledge in a graph; (2) We incorporate the prior knowledge with graph-based knowledge inference to enhance report generation; (3) Extensive experiments on the IU X-ray dataset \cite{demner-fushman_preparing_2016} show that our proposed model outperforms state-of-the-art methods. (4) We make codes, models, and pre-processed data publicly available.

The rest of the paper is organized as follows. We describe the multi-task learning in Section~\ref{sec:methods}, followed by our experimental setup, results, and discussion in Section~\ref{sec:results}. We conclude with future work in the last section.

\section{Methods}
\label{sec:methods}

\subsection{Framework}
\begin{figure}[h]
    \centering
    \includegraphics[width=\columnwidth,page=3,clip,trim=0 31em 2em 0]{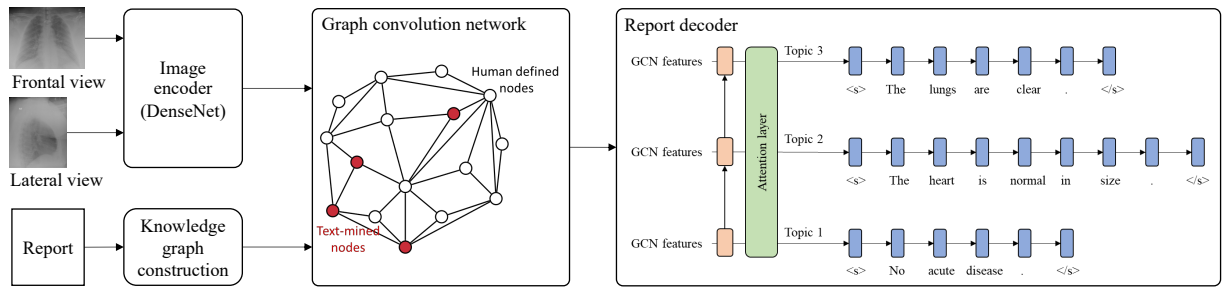}
    \caption{The proposed framework.}
    \label{fig:overview}
\end{figure}

Simulating the reading routine of radiologists by first observing multiple findings when they read medical images and then compiling radiological reports, our proposed method generates radiology report $S$ from the frontal-view $I_f$ and lateral-view $I_l$ of chest X-ray images following several steps (Fig~\ref{fig:overview}). We first constructed the The prior knowledge graph (i.e., relationships between medical findings) in a data-driven manner (Section~\ref{knowledge}). The frontal-view and lateral-view images are fed to an image encoder to extract visual features (Section~\ref{sec:encoder}), which are then passed to a graph convolution network (GCN) based on the knowledge graph (Section~\ref{sec:gcn}). The network then propagates the semantic correlations among radiology concepts based on the prior knowledge graph, therefore, injecting domain knowledge into concept representation learning. We then concatenate the node features of both views and train a report decoder (Section~\ref{sec:decoder}).


\subsection{Prior Knowledge Graph Construction}
\label{knowledge}

In our study, the nodes in the knowledge graph are radiology concepts (e.g., diseases or body parts) and edges are the semantic correlations among the concepts. Our knowledge graph consists of two parts. The first part was manually defined by domain experts\cite{zhang_when_2020}. The second part consists of supplementary concepts and their correlations text-mined from the radiology reports in a data-driven manner. Specifically, we first build a rule-based tool to greedily match concepts in RadLex on sequences of the lemmatized tokens in the reports (i.e., longer matches are returned where possible). We then select the concepts with top-$q$ appearing frequencies if they have not been included in the graph in Zhang et al.\cite{zhang_when_2020}. Here, we only focus on three categories of interest: Anatomical entity, Clinical finding, and Imaging observation. We then examine the document-level co-occurrences of concepts to build a correlation matrix and binarize the matrix to prevent overfitting\cite{chen2019multi}.

\subsection{Image Encoder}
\label{sec:encoder}

We employ DenseNet-121 \cite{huang_added_2018} (pre-trained on CheXpert\cite{irvin_chexpert_2019}) as our image encoder backbone. One frontal-view and one lateral-view chest X-ray image are fed into DenseNet-121 to extract visual features, which are then used to initialize graph node features in two steps. The first step initializes the feature of the global node by average pooling the visual features of the frontal-view and lateral-view images. The second step initializes the features of the remaining finding nodes in the graph through a spatial attention mechanism. We use a convolution layer with kernel size one followed by a softmax to compute the spatial attention weights, where the number of channels is equal to the number of finding nodes in the graph. Finally, we concatenate the global node feature computed from step one, with the weighted sum of visual features, where the weights come from step two, to initialize the graph node features.

\subsection{Graph Convolution Network}
\label{sec:gcn}

We employ a GCN to model inner correlations among radiology concepts. The graph structure is constructed based on the graph detailed above. The GCN updates its node representations by message passing. The graph convolution is expressed as\cite{Kipf-2017-1}:
\begin{align*}
    \hat{H}^l & = ReLU(BN(Conv1d(H^{l})))\\
    m & = ReLU(D^{-1/2}\hat{A}D^{-1/2}H^{l}W^{l})\\
    H^{l+1} & = ReLU(BN(Conv1d(concat(\hat{H}^{l}, m)))
\end{align*}
where $H^{l}$ is the states in the $l$-th layer, with $H^0$ initialized using the output of image encoder. $\hat{A}$ = $A$ + $I_{N}$ is the adjacency matrix with added self-connections, where $A$ is the graph adjacency matrix, $I_{N}$ is the $N$-dimension identity matrix, $D=\text{diag}\sum_{j}^{} A_{ij}$ is the diagonal node degree matrix, $BN$ is the batch normalization, and $W^{l}$ is a trainable layer-specific weight matrix.

\subsection{Report Generation Decoder}
\label{sec:decoder}

Radiology reports usually contain several sentences where each sentence focuses on one topic. Therefore, we adopt a two-level LSTM structure\cite{zhang_when_2020}. We input the graph node features to an attention module to obtain the context vector which attends graph node features to different topics. The vector is then fed to a topic-LSTM to generate topics, and the output topic vectors are passed to a word-LSTM to generate sentences in a word-by-word fashion.

\subsection{Training Procedure and Loss Functions}
\label{sec-classification}
Our framework combines two loss functions. Suppose $p(S_t)$ is the probability of observing the correct word $S_t$ at time $t$. The first loss is the cross-entropy loss of the predictions on the whole report, 
\begin{align*}
    -\sum_{t=1}^{N} \log p (S_t|I_l,I_f;\theta)
\end{align*}

Additionally, we take the features of each node in the graph and do average pooling. We then fit a linear classifier to predict the diseases present in the images. 
Clinically, it simulates the reading routine of radiologists by first observing multiple findings when they read medical images and then compiling radiology reports. 

To use the ground truth labels, we divide the nodes in the knowledge graph into two types: primary and auxiliary. The primary nodes are the chest observations with ground truth labels in the dataset. The auxiliary nodes are the supplementary concepts mined from the reports. We then use weighted binary cross-entropy loss defined on the primary nodes \cite{Wang-2017} for training. 

Our model is trained using the same two-step training procedure as in Zheng et al. \cite{zhang_when_2020}: the multi-label classifier is trained first, and then we fixed the image encoder and GCN modules when training the report decoder.

\section{Results}
\label{sec:results}
\subsection{Datasets}

\begin{wrapfigure}{r}{0.40\textwidth}
\captionof{table}{\label{settings} Dataset statistics.}
\centering
\begin{tabular}{lrrr}
\toprule
 & Reports & Sentences & Tokens \\
\midrule
Training &  1,747 & 10,966 & 80,491 \\
Validation &  582 & 3,641 & 27,048 \\
Test &  583 & 3,659 & 27,113 \\
\bottomrule
\end{tabular}
\end{wrapfigure}
We use the publicly available IU X-ray dataset \cite{demner-fushman_preparing_2016}. IU X-ray dataset contains 3,955 de-identified radiology reports, with each report associated with one frontal-view, and one lateral-view chest X-ray image. Several sections (e.g., findings and impressions) are covered in each radiology report, where findings describe the medical findings and impressions summarize the overall diagnoses. 

We only consider the cases with complete findings and impression sections, and with both frontal-view and lateral-view images present, which results in 2,912 reports and 5,824 images. We concatenate the findings and impression sections in each report as the ground truth. All the words in the ground truth are tokenized, converted to lower case. Infrequent words with a frequency of less than three are dropped, which results in 1,103 unique tokens. The ground truth labels are obtained by detecting the corresponding labels in the Mesh part of the reports, where findings are listed in a formatted manner. 
After pre-processing, we have a total of 2,912 reports and 5,824 images (Table~\ref{settings}), where each report is associated with one frontal-view and one lateral-view of the chest X-ray image. 

\subsection{Experimental Settings}

We adopt DenseNet-121\cite{huang_added_2018} pre-trained on CheXpert\cite{irvin_chexpert_2019} as our backbone CNN model. Images are randomly cropped to 512$\times$512 with padding if needed, and the feature map from DenseNet-121 block four is of size 1024$\times$16$\times$16. We replace the last fully connected layer of DenseNet-121 with a multi-label classification layer appended with attention and three graph convolution layers with 256 hidden units. 

Our model is trained for 150 epochs with a batch size of 8. Adam \cite{kingma_adam_2015} is used for optimization with a learning rate of 1e-6 and weight decay of 1e-5. We use 200-dimension GloVe word embeddings \cite{glove-2014} in the decoder. 

\subsection{Evaluation metrics}

To evaluate the generation results, we compute BLEU scores \cite{bleu-2002}, ROUGE-L \cite{lin_rouge_2004}, and CIDEr \cite{cider-2015}, which reflect different aspects of the performance, where BLEUs reflect the precision, ROUGE-L is closer to recall, and CIDEr measures consensus. 

In this study, we used the bootstrap to assess the statistical significance of the results. For the training dataset, we sampled 1,747 instances with replacement to train the models. We then evaluated the model on the held-out test dataset. By repeating this sampling, training, and evaluation 15 times, we obtained a distribution of the performance metrics and reported the 95\% confidence intervals (CI).

\subsection{Results and Discussion}
\subsubsection{Report generation} 
We compare our method with four previous works on radiology report generation, including SAT framework in Xu et al. \cite{xu_show_2015}, multi-level LSTM framework in Yuan et al.\cite{yuan-2019}, two-level LSTM augmented with knowledge graph in Zhang et al.\cite{zhang_when_2020}, and CoAtt in Jing et al. \cite{jing-2018}. Table~\ref{single-column-result-1} shows the performance comparison. Our method achieves 0.4\%-2.0\% improvements over previous works, especially 2.0\% improvement in CIDEr, 1.5\% in ROUGE-L, and 1.0\% in BLEU-2. 

\begin{table*}[h]
\small
\vspace{1em}
\centering
\caption{\label{single-column-result-1}Comparisons on the test set of IU X-ray dataset. The 95\% CI is reported for our model.}
\begin{tabular*}{\textwidth}{lccccccc}
\toprule
                                    & BLEU-1 & BLEU-2 & BLEU-3 & BLEU-4 & ROUGE-L & CIDEr & \textit{Mean}  \\
\midrule
Xu et al. \cite{xu_show_2015}               & 0.433  & 0.281  & 0.194  & 0.138  & 0.361 & 0.320 & 0.288 \\
Yuan et al. \cite{yuan-2019}             & 0.445  & 0.289  & 0.200  & 0.143  & 0.359 & 0.268 & 0.284 \\

Jing et al. \cite{jing-2018}          & \textbf{0.455}  & 0.288  & 0.205  & 0.154  & 0.369 & 0.277 & 0.291 \\

Zhang et al. \cite{zhang_when_2020}            & 0.441  & 0.291  & 0.203  & 0.147  & 0.367 & 0.304 & 0.292 \\

Our model                           & 0.450$\pm$0.005  & \textbf{0.301$\pm$0.003}  & \textbf{0.213$\pm$0.004} & \textbf{0.158$\pm$0.005}  & \textbf{0.384$\pm$0.007} & \textbf{0.340$\pm$0.011} & 
\textbf{0.308} \\
\bottomrule
\end{tabular*}
\vspace{.5em}
\vspace{-.5em}
\end{table*}

\subsubsection{Ablation studies}

We conducted ablation studies to verify the effectiveness of different modules (Table~\ref{single-column-result-2}). According to (a), adopting the pre-trained word embeddings (GloVe) improves all metrics, where the average score increases by 1.4\%. When the proposed GCN is added (ablation study (b)), the BLEU scores increase by up to 0.6\% and ROUGE-L increases by 0.6\%. Note that CIDEr drops probably because the generated reports contained repeated sentences due to a lack of contextual coherence;
while compared to other metrics, CIDEr intends to weigh more on the details of sentences.

\begin{table}[h]
\vspace{1em}
\centering
\caption{\label{single-column-result-2}Ablation studies on the test set of IU X-ray dataset.}
\begin{tabular}{lccccccc}
\toprule
                                    & BLEU-1 & BLEU-2 & BLEU-3 & BLEU-4 & ROUGE-L & CIDEr & \textit{Mean}  \\
\midrule
Our model                           & 0.450  & 0.301  & 0.213  & 0.158  & 0.384 & 0.340 & 0.308 \\
\midrule
(a) Random embs                     & 0.421  & 0.281  & 0.202  & 0.151  & 0.367 & 0.340 & 0.294 \\
(b) GCN in Zhang et al. \cite{zhang_when_2020} & 0.446  & 0.297  & 0.209  & 0.152  & 0.378 & \textbf{0.357} & 0.307 \\
(c) 20 Nodes                        & \textbf{0.460}  & 0.307  & 0.215  & 0.155  & 0.380 & 0.287 & 0.301 \\
(d) 40 Nodes                        & \textbf{0.460}  & \textbf{0.311}  & \textbf{0.223}  & \textbf{0.164}  & \textbf{0.387} & 0.310 & \textbf{0.309} \\
(e) 60 Nodes                        & 0.446  & 0.296  & 0.209  & 0.152  & 0.372 & 0.318 & 0.299 \\
\bottomrule
\end{tabular}
\end{table}

We also compare different sizes of nodes in GCN. In addition to the nodes in Zhang et al. \cite{zhang_when_2020}, we text-mined ten auxiliary nodes from the IU X-ray dataset, including Chest pain, Shortness of breath, Granuloma, Lymphadenopathy, Deformity, Granulomatous disease, Congestion, Tuberculosis, Infection, and Hypertension. Ablation studies (c) - (e) show that the performance drops significantly when switching to a larger graph with 60 nodes. This observation indicates that a larger graph may not always lead to more accurate report generation. 

\subsubsection{Error analysis}

\begin{wrapfigure}{r}{0.45\columnwidth}
    \centering
    \vspace{-2em}
    \includegraphics[width=.45\columnwidth]{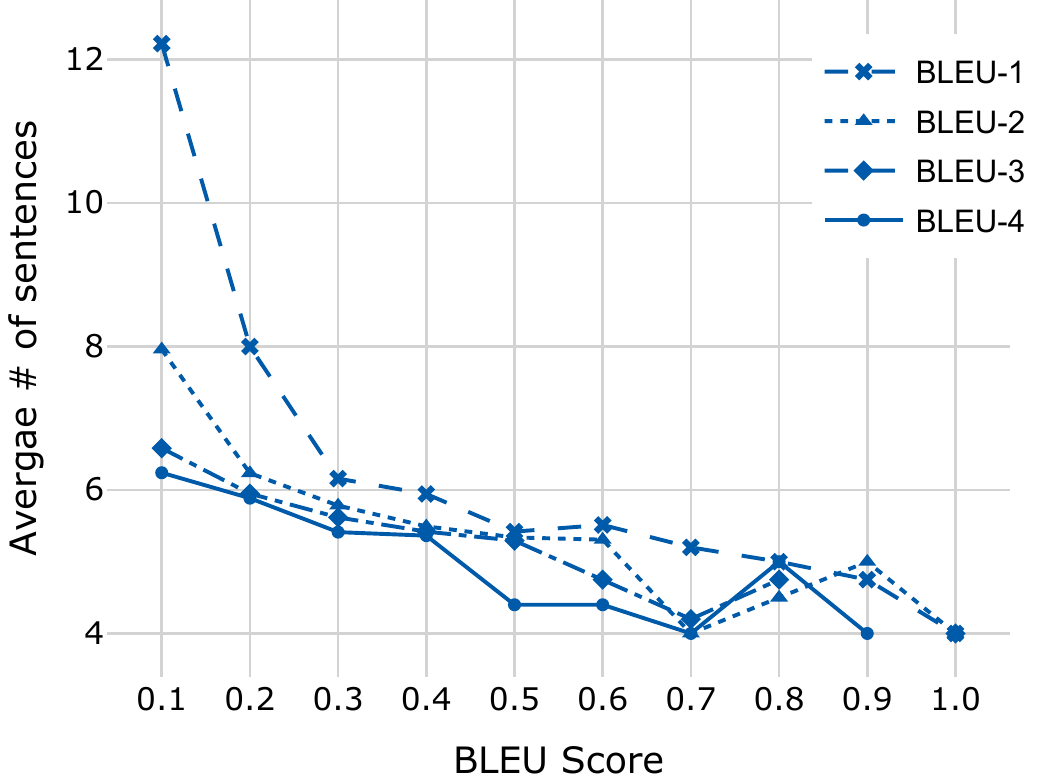}
    \caption{Average sentence number of reports of different BLEU scores.}
    \label{number-sentence}
    \vspace{1em}
\end{wrapfigure}
We further studied the correlations between text complexity and evaluation metrics. Fig~\ref{number-sentence} shows the average number of sentences in the report for different BLEU scores. We find that reports with more sentences are prone to be less accurate and tend to have lower BLEU scores.

\subsubsection{Multi-label classification}

We monitored the node features in the GCN and measured how suitable they are for the disease classification. We apply global average pooling after the graph convolution layers to obtain the graph-level feature, and further, append a fully connected layer to predict probabilities for each finding as a multi-label classification task. We use Binary Cross Entropy loss during training. While we are using 30 knowledge graph nodes, we only classify on the first 20 finding nodes, leaving the rest 10 finding nodes as auxiliary nodes and not including them when calculating the loss. The numbers of reports labeled with different diseases are as shown in Table \ref{mlclass-report-numbers}.

\begin{table}[h]
\vspace{1em}
\centering
\caption{\label{mlclass-report-numbers} Disease report distributions, classification accuracies and the quality of generated reports.}
\vspace{-1em}
\begin{tabular}{lrrrrrrrr}
\toprule
Disease & \multicolumn{2}{c}{Reports} & \multicolumn{2}{c}{AUC} & BLEU-1 & BLEU-2 & BLEU-3 & BLEU-4\\
& & \% & Zhang et al.\cite{zhang_when_2020} & Proposed\\
\midrule
Normal & 1,491 & 38 & 0.81 & 0.81 & 0.47 & 0.33 & 0.24 & 0.16\\
Airspace disease & 125 & 3 & 0.86 & 0.81 & 0.31 & 0.18 & 0.10 & 0.05\\
Atelectasis & 332 & 8 & 0.67 & 0.70 & 0.41 & 0.27 & 0.19 & 0.13\\
Calcinosis & 305 & 8 & 0.91 & 0.91 & 0.30 & 0.18 & 0.11 & 0.06\\
Cardiomegaly & 375 & 10 & 0.73 & 0.79 & 0.32 & 0.19 & 0.11 & 0.05\\
Cicatrix & 196 & 5 & 0.89 & 0.94 & 0.25 & 0.15 & 0.09 & 0.04\\
Edema & 46 & 1 & 0.67 & 0.76 & 0.36 & 0.18 & 0.07 & 0.05\\
Effusion & 161 & 4 & 0.88 & 0.69 & 0.32 & 0.18 & 0.10 & 0.06\\
Emphysema & 106 & 3 & 0.78 & 0.79 & 0.37 & 0.24 & 0.15 & 0.09\\
Fracture bone & 84 & 2 & 0.94 & 0.97 & 0.34 & 0.22 & 0.15 & 0.10\\
Hernia & 48 & 1 & 0.81 & 0.80 & 0.29 & 0.18 & 0.10 & 0.07\\
Hypoinflation & 507 & 13 & 0.64 & 0.60 & 0.27 & 0.16 & 0.09 & 0.05\\
Lesion & 126 & 3 & 0.80 & 0.82 & 0.35 & 0.22 & 0.14 & 0.09\\
Medical device & 362 & 9 & 0.86 & 0.83 & 0.37 & 0.25 & 0.18 & 0.12\\
Opacity & 455 & 12 & 0.84 & 0.77 & 0.30 & 0.19 & 0.11 & 0.07\\
Pneumonia & 120 & 3 & 0.83 & 0.83 & 0.28 & 0.18 & 0.12 & 0.08\\
Pneumothorax & 27 & 1 & 0.93 & 0.87 & 0.29 & 0.17 & 0.09 & 0.05\\
Scoliosis & 559 & 14 & 0.66 & 0.64 & 0.37 & 0.24 & 0.15 & 0.08\\
Thickening & 56 & 1 & 0.73 & 0.77 & 0.34 & 0.23 & 0.15 & 0.10\\
Others & 411 & 10 & 0.60 & 0.61 & 0.44 & 0.30 & 0.21 & 0.14\\
\bottomrule
\end{tabular}
\vspace{.5em}
\vspace{-1.5em}
\end{table}




Table \ref{mlclass-report-numbers} also shows that our method achieves comparable results in AUC with the method described in Zhang et al.\cite{zhang_when_2020}. For diseases that appear in more than 10\% of the reports (Scoliosis, Hypoinflatin, Opacity, and Cardiomedgaly), the correlation between AUC and BLEU is moderate (0.46, 0.44, 0.45, 0.44, respectively). For other diseases, there are no dependencies between disease classifications and BLEU scores. For example, pleural effusion has a better classification accuracy than Emphysema, but this does not suggest that pleural effusion will have a better report generation accuracy. However, since the number of reports per disease is imbalanced in the IU dataset, more datasets are needed to further investigate their correlations.

\subsubsection{Error case analysis}
\begin{figure}[h!]
    \centering
    \includegraphics[width=\columnwidth,page=1,clip]{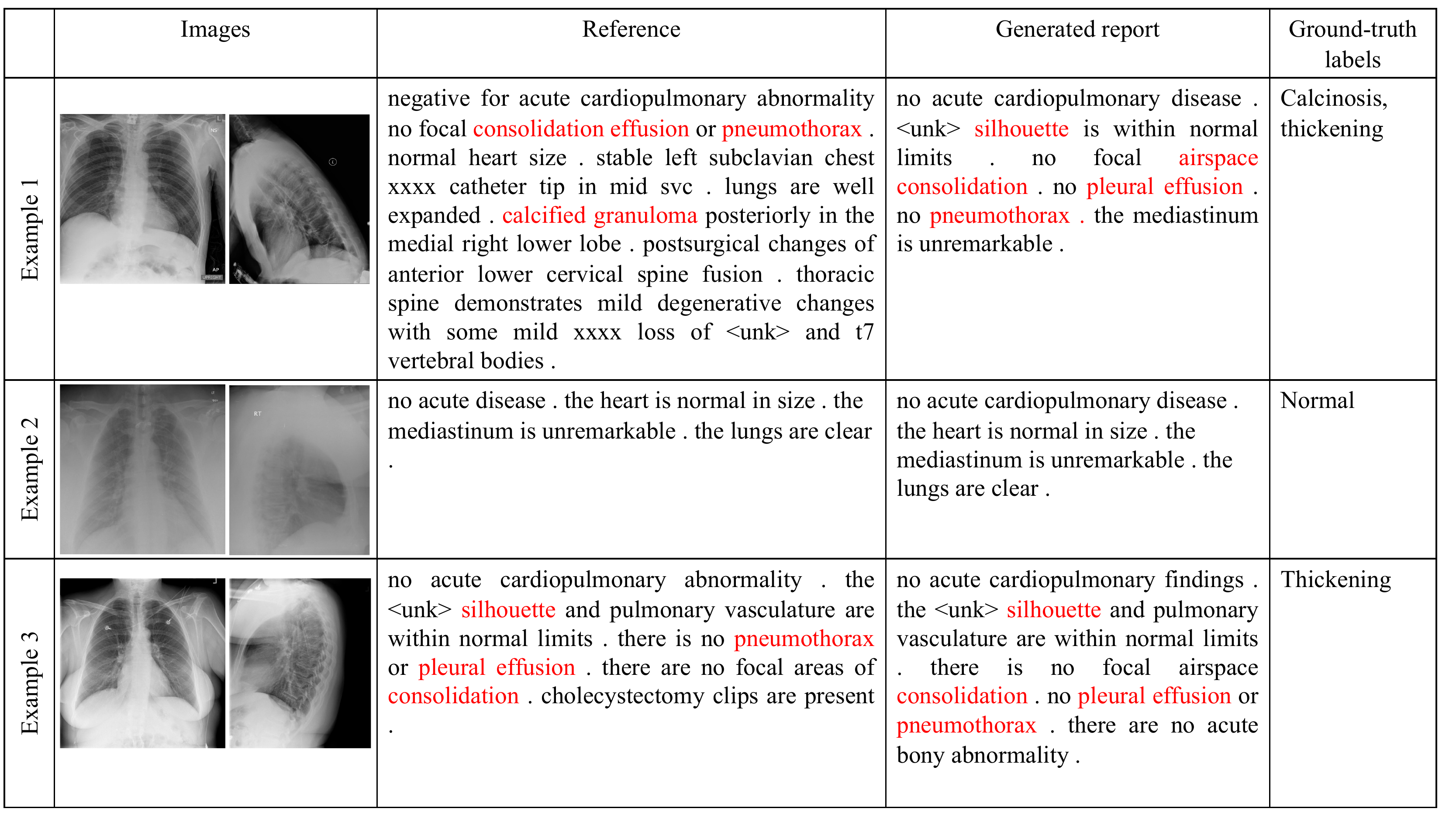}
    \caption{Three examples where our model generates low-scored reports and high-scored reports.}
    \label{tab:error-analysis}
    \vspace{.5em}
\end{figure}

Fig \ref{tab:error-analysis} shows three examples where our model respectively generates a low-scored report and two high-scored reports. The first example is labeled with Calcinosis and Thickening. However, our generated report omits the fact that there is a calcified granuloma. The BLEU-1 score is 0.1389. Example 2 and Example 3 are respectively labeled with Normal and Thickening. We can see that the generated reports are very close to the ground truth reports in terms of concept mentions and descriptions. They report BLEU-1 scores 0.9546 and 0.6757, respectively.

\subsubsection{Limitations and Discussions}
Our framework employs DenseNet-121 pre-trained on CheXpert as our backbone, which has the underlying limitation that CheXNet\cite{rajpurkar2017chexnet} was fine-tuned on one single-institution dataset.  Chest X-ray image quality can also affect the training performance. To understand the impact of image acquisition quality, we experimented adding random Gaussian noises to the training images, and the average multi-label classification AUC drops from 0.786 to 0.683. With the six report generation metrics dropping to 0.384, 0.271, 0.197, 0.145, 0.394, 0.301 respectively, the average accuracy of the generated reports drops from 0.308 to 0.282.

\section{Conclusions}

In this paper, we propose to incorporate text-mined prior knowledge with radiology reporting by employing a graph convolution module for knowledge inference, followed by multi-label disease classification and report generation. Our model achieves better performances than previous works on the IU X-ray dataset. Furthermore, we verified the effectiveness of different modules through ablation studies. 

In the future, we plan to adopt Transformer to improve the contextual coherence and explore other domain knowledge that can be utilized in report generation. We plan to train and evaluate our model on different datasets. While our work only scratches the surface, we hope it will shed light on the future directions for radiology reporting.

\section*{Acknowledgment}

This work is supported by Amazon Machine Learning Research Award 2020. It was also supported by the National Library of Medicine under Award No. 4R00LM013001.

\makeatletter
\renewcommand{\@biblabel}[1]{\hfill #1.}
\makeatother

\bibliography{bibm_bib,zetero_ref}

\begin{thebibliography}{10}

\bibitem{pahadia_khurana_geha_deahl_2020}
Pahadia M, Khurana S, Geha H, Deahl STI.
\newblock Radiology report writing skills: A linguistic and technical guide for
  early-career oral and maxillofacial radiologists.
\newblock Imaging Science in Dentistry. 2020 Sep;50(3):269–272.

\bibitem{shin2015interleaved}
Shin HC, Lu L, Kim L, Seff A, Yao J, Summers RM.
\newblock Interleaved text/image deep mining on a large-scale radiology
  database for automated image interpretation.
\newblock The Journal of Machine Learning Research. 2016;17(1):3729--3759.

\bibitem{rennie2017selfcritical}
Rennie SJ, Marcheret E, Mroueh Y, Ross J, Goel V.
\newblock Self-Critical Sequence Training for Image Captioning.
\newblock In: 2017 IEEE Conference on Computer Vision and Pattern Recognition
  (CVPR). Los Alamitos, CA, USA: IEEE Computer Society; 2017. p. 1179--1195.
\newblock Available from:
  \url{https://doi.ieeecomputersociety.org/10.1109/CVPR.2017.131}.

\bibitem{lu2017knowing}
Lu J, Xiong C, Parikh D, Socher R.
\newblock Knowing When to Look: Adaptive Attention via a Visual Sentinel for
  Image Captioning.
\newblock In: 2017 IEEE Conference on Computer Vision and Pattern Recognition
  (CVPR); 2017. p. 3242--3250.

\bibitem{anderson2018bottomup}
Anderson P, He X, Buehler C, Teney D, Johnson M, Gould S, et~al.
\newblock Bottom-Up and Top-Down Attention for Image Captioning and Visual
  Question Answering.
\newblock In: 2018 IEEE/CVF Conference on Computer Vision and Pattern
  Recognition; 2018. p. 6077--6086.

\bibitem{agrawal2016vqa}
Antol S, Agrawal A, Lu J, Mitchell M, Batra D, Zitnick CL, et~al.
\newblock VQA: Visual Question Answering.
\newblock In: 2015 IEEE International Conference on Computer Vision (ICCV);
  2015. p. 2425--2433.

\bibitem{jing-2018}
Jing B, Xie P, Xing E.
\newblock On the Automatic Generation of Medical Imaging Reports.
\newblock In: Proceedings of the 56th Annual Meeting of the Association for
  Computational Linguistics (Volume 1: Long Papers). Melbourne, Australia:
  Association for Computational Linguistics; 2018. p. 2577--2586.
\newblock Available from: \url{https://aclanthology.org/P18-1240}.

\bibitem{wang2018tienet}
Wang X, Peng Y, Lu L, Lu Z, Summers RM.
\newblock TieNet: Text-Image Embedding Network for Common Thorax Disease
  Classification and Reporting in Chest X-Rays.
\newblock In: 2018 IEEE/CVF Conference on Computer Vision and Pattern
  Recognition (CVPR); 2018. p. 9049--9058.

\bibitem{jing-etal-2019-show}
Jing B, Wang Z, Xing E.
\newblock Show, Describe and Conclude: On Exploiting the Structure Information
  of Chest {X}-ray Reports.
\newblock In: Proceedings of the 57th Annual Meeting of the Association for
  Computational Linguistics. Florence, Italy: Association for Computational
  Linguistics; 2019. p. 6570--6580.
\newblock Available from: \url{https://www.aclweb.org/anthology/P19-1657}.

\bibitem{chen-etal-2020-generating-radiology}
Chen Z, Song Y, Chang TH, Wan X.
\newblock Generating Radiology Reports via Memory-driven Transformer.
\newblock In: Proceedings of the 2020 Conference on Empirical Methods in
  Natural Language Processing (EMNLP). Online: Association for Computational
  Linguistics; 2020. p. 1439--1449.
\newblock Available from:
  \url{https://www.aclweb.org/anthology/2020.emnlp-main.112}.

\bibitem{xue2018}
Xue Y, Xu T, {Rodney Long} L, Xue Z, Antani S, Thoma G, et~al.
\newblock Multimodal recurrent model with attention for automated radiology
  report generation.
\newblock In: Medical Image Computing and Computer Assisted Intervention –
  MICCAI 2018. Germany; 2018. p. 457--466.

\bibitem{liu2019clinically}
Liu G, Hsu TMH, McDermott M, Boag W, Weng WH, Szolovits P, et~al.
\newblock Clinically accurate chest x-ray report generation.
\newblock In: Machine Learning for Healthcare Conference. PMLR; 2019. p.
  249--269.

\bibitem{zhang_when_2020}
Zhang Y, Wang X, Xu Z, Yu Q, Yuille A, Xu D.
\newblock When {Radiology} {Report} {Generation} {Meets} {Knowledge} {Graph}.
\newblock Proceedings of the AAAI Conference on Artificial Intelligence. 2020
  Apr;34(07):12910--12917.
\newblock Available from:
  \url{https://aaai.org/ojs/index.php/AAAI/article/view/6989}.

\bibitem{li2019knowledgedriven}
Li CY, Liang X, Hu Z, Xing EP.
\newblock Knowledge-driven encode, retrieve, paraphrase for medical image
  report generation.
\newblock Proceedings of the AAAI Conference on Artificial Intelligence. 2019
  Jul;33:6666--6673.

\bibitem{langlotz_radlex_2006}
Langlotz CP.
\newblock {RadLex}: a new method for indexing online educational materials.
\newblock Radiographics: A Review Publication of the Radiological Society of
  North America, Inc. 2006 Dec;26(6):1595--1597.

\bibitem{yao_exploring_2018}
Yao T, Pan Y, Li Y, Mei T.
\newblock Exploring {Visual} {Relationship} for {Image} {Captioning}.
\newblock In: Ferrari V, Hebert M, Sminchisescu C, Weiss Y, editors. Computer
  {Vision} – {ECCV} 2018. Lecture {Notes} in {Computer} {Science}. Cham:
  Springer International Publishing; 2018. p. 711--727.

\bibitem{HochSchm-1997}
Hochreiter S, Schmidhuber J.
\newblock {Long Short-Term Memory}.
\newblock Neural Computation. 1997 11;9(8):1735--1780.
\newblock Available from: \url{https://doi.org/10.1162/neco.1997.9.8.1735}.

\bibitem{demner-fushman_preparing_2016}
Demner-Fushman D, Kohli MD, Rosenman MB, Shooshan SE, Rodriguez L, Antani S,
  et~al.
\newblock Preparing a collection of radiology examinations for distribution and
  retrieval.
\newblock Journal of the American Medical Informatics Association : JAMIA. 2016
  Mar;23(2):304--310.

\bibitem{chen2019multi}
Chen ZM, Wei XS, Wang P, Guo Y.
\newblock Multi-Label Image Recognition With Graph Convolutional Networks.
\newblock In: 2019 IEEE/CVF Conference on Computer Vision and Pattern
  Recognition (CVPR); 2019. p. 5172--5181.

\bibitem{huang_added_2018}
Huang P, Park S, Yan R, Lee J, Chu LC, Lin CT, et~al.
\newblock Added value of computer-aided {CT} image features for early lung
  cancer diagnosis with small pulmonary nodules: {A} matched case-control
  study.
\newblock Radiology. 2018 Jan;286(1):286--295.

\bibitem{irvin_chexpert_2019}
Irvin J, Rajpurkar P, Ko M, Yu Y, Ciurea-Ilcus S, Chute C, et~al.
\newblock {CheXpert}: a large chest radiograph dataset with uncertainty labels
  and expert comparison.
\newblock In: Proceedings of the {AAAI} {Conference} on {Artificial}
  {Intelligence}. vol.~33; 2019. p. 590--597.

\bibitem{Kipf-2017-1}
Jia N, Tian X, Zhang Y, Wang F.
\newblock Semi-Supervised Node Classification With Discriminable Squeeze
  Excitation Graph Convolutional Networks.
\newblock IEEE Access. 2020;8:148226--148236.

\bibitem{Wang-2017}
Wang X, Peng Y, Lu L, Lu Z, Bagheri M, Summers RM.
\newblock ChestX-Ray8: Hospital-Scale Chest X-Ray Database and Benchmarks on
  Weakly-Supervised Classification and Localization of Common Thorax Diseases.
\newblock 2017 IEEE Conference on Computer Vision and Pattern Recognition
  (CVPR). 2017 Jul.
\newblock Available from: \url{http://dx.doi.org/10.1109/CVPR.2017.369}.

\bibitem{kingma_adam_2015}
Kingma DP, Ba J.
\newblock Adam: a method for stochastic optimization.
\newblock In: International {Conference} on {Learning} {Representations}
  ({ICLR}); 2015. p. 1--15.
\newblock Available from: \url{https://arxiv.org/abs/1412.6980}.

\bibitem{glove-2014}
Pennington J, Socher R, Manning C.
\newblock {G}lo{V}e: Global Vectors for Word Representation.
\newblock In: Proceedings of the 2014 Conference on Empirical Methods in
  Natural Language Processing ({EMNLP}). Doha, Qatar: Association for
  Computational Linguistics; 2014. p. 1532--1543.
\newblock Available from: \url{https://www.aclweb.org/anthology/D14-1162}.

\bibitem{bleu-2002}
Papineni K, Roukos S, Ward T, Zhu WJ.
\newblock {B}leu: a Method for Automatic Evaluation of Machine Translation.
\newblock In: Proceedings of the 40th Annual Meeting of the Association for
  Computational Linguistics. Philadelphia, Pennsylvania, USA: Association for
  Computational Linguistics; 2002. p. 311--318.
\newblock Available from: \url{https://www.aclweb.org/anthology/P02-1040}.

\bibitem{lin_rouge_2004}
Lin CY.
\newblock {ROUGE}: a package for automatic evaluation of summaries.
\newblock In: Text summarization branches out: {Proceedings} of the {ACL}-04
  workshop. vol.~8. Barcelona, Spain; 2004. p. 1--8.
\newblock Available from: \url{https://aclanthology.org/W04-1013/}.

\bibitem{cider-2015}
Vedantam R, Zitnick CL, Parikh D.
\newblock {{CIDEr}}: {{Consensus}}-based image description evaluation.
\newblock In: 2015 {{IEEE Conference}} on {{Computer Vision}} and {{Pattern
  Recognition}} ({{CVPR}}). {Boston, MA, USA}: {IEEE}; 2015. p. 4566--4575.

\bibitem{xu_show_2015}
Xu K, Ba J, Kiros R, Cho K, Courville A, Salakhudinov R, et~al.
\newblock Show, attend and tell: neural image caption generation with visual
  attention.
\newblock In: International {Conference} on {Machine} {Learning} ({ICML});
  2015. p. 2048--2057.
\newblock Available from: \url{https://arxiv.org/abs/1502.03044}.

\bibitem{yuan-2019}
Yuan J, Liao H, Luo R, Luo J.
\newblock Automatic radiology report generation based on multi-view image
  fusion and medical concept enrichment.
\newblock In: Shen D, Liu T, Peters TM, Staib LH, Essert C, Zhou S, et~al.,
  editors. Medical {{Image Computing}} and {{Computer Assisted Intervention}}
  \textendash{} {{MICCAI}} 2019. vol. 11769. {Cham}: {Springer International
  Publishing}; 2019. p. 721--729.

\bibitem{rajpurkar2017chexnet}
Rajpurkar P, Irvin J, Zhu K, Yang B, Mehta H, Duan T, et~al.. CheXNet:
  Radiologist-Level Pneumonia Detection on Chest X-Rays with Deep Learning;
  2017.

\end{thebibliography}
\bibliographystyle{vancouver}

\end{document}